\documentclass{article}

\usepackage[preprint]{neurips_2025}

\usepackage[utf8]{inputenc} %
\usepackage[T1]{fontenc}    %
\usepackage{hyperref}       %
\usepackage{url}            %
\usepackage{booktabs}       %
\usepackage{amsfonts}       %
\usepackage{nicefrac}       %
\usepackage{microtype}      %
\usepackage{xcolor}         %

\usepackage{enumitem} 
\usepackage{amsmath}
\usepackage{amsthm}    %
\usepackage[most]{tcolorbox}
\usepackage{lineno}
\usepackage{tikz}
\usepackage{fontawesome5}
\usepackage{svg}
\usepackage{multirow}
\usepackage{amsfonts}       %
\usepackage{array}

\usepackage{algorithm}
\usepackage{algpseudocode}

\newtheorem{theorem}{Theorem}
\title{CoTGuard: Using Chain-of-Thought Triggering for Copyright Protection in Multi-Agent LLM Systems}

\author{%
  Yan Wen \\
  Department of Computer Science\\
  University of Maryland, College Park \\
  College Park, MD 20742 \\
  \texttt{ywen1@umd.edu} \\
  \And
  Junfeng Guo \\
  Department of Computer Science\\
  University of Maryland, College Park \\
  College Park, MD 20742 \\
  \texttt{gjf2023@umd.edu} \\
  \And
    Heng Huang \\
  Department of Computer Science\\
  University of Maryland, College Park \\
  College Park, MD 20742 \\
  \texttt{heng@umd.edu} \\
}

\begin{document}

\maketitle

\begin{abstract}

  As large language models (LLMs) evolve into autonomous agents capable of collaborative reasoning and task execution, multi-agent LLM systems have emerged as a powerful paradigm for solving complex problems. However, these systems pose new challenges for copyright protection, particularly when sensitive or copyrighted content is inadvertently recalled through inter-agent communication and reasoning. Existing protection techniques primarily focus on detecting content in final outputs, overlooking the richer, more revealing reasoning processes within the agents themselves. In this paper, we introduce CoTGuard, a novel framework for copyright protection that leverages trigger-based detection within Chain-of-Thought (CoT) reasoning. Specifically, we can activate specific CoT segments and monitor intermediate reasoning steps for unauthorized content reproduction by embedding specific trigger queries into agent prompts. This approach enables fine-grained, interpretable detection of copyright violations in collaborative agent scenarios. We evaluate CoTGuard on various benchmarks in extensive experiments and show that it effectively uncovers content leakage with minimal interference to task performance. Our findings suggest that reasoning-level monitoring offers a promising direction for safeguarding intellectual property in LLM-based agent systems.
\end{abstract}

\section{Introduction}
Recent advances in large language models (LLMs), such as GPT-4 \cite{achiam2023gpt}, Genimi \cite{team2023gemini},  DeepSeek \cite{guo2025deepseek}, have significantly transformed natural language processing (NLP), enabling a wide array of applications across writing \cite{yuan2022wordcraft}, translation \cite{zhang2023prompting}, coding \cite{nijkamp2022codegen}, and reasoning \cite{plaat2024reasoning}. Building on the generalization and zero-shot capabilities of LLMs, researchers have developed LLM-based agent systems \cite{li2024survey} that simulate autonomous agents capable of planning \ cite {xie2024travelplanner}, collaboration \cite{liu2023dynamic}, and task execution \cite{park2023generative}. These multi-agent systems leverage LLMs as their core reasoning engines, often coordinating via natural language to achieve complex objectives, from web automation to collaborative problem-solving.

However, the rise of LLMs and their deployment in agent-based systems has introduced pressing concerns about intellectual property and copyright protection \cite{ren2024copyright, chu2024protect}. Much current research in LLM-related copyright protection focuses on detecting memorization or leakage of training data, watermarking generated content,   and legal frameworks for model training on copyrighted corpora \cite{guo2023domain, li2024double, wang2024espew, xu2025can,liu2024shield}. However, relatively little work has extended these protections to LLM-based agent systems, where models interact in more complex, emergent behaviors that make unauthorized content reproduction more challenging to trace \cite{bender2021dangers, park2023generativeagents, xu2024adversarial}. While research on single-agent LLM copyright protection is well-established \cite{carlini2023extracting, zou2023unlearnable}, multi-agent settings introduce unique challenges due to the collaborative, distributed nature of the reasoning process \cite{guo2023coda}.

Motivated by Chain-of-Thought (CoT) reasoning \cite{wei2022chain}, we identify a novel attack surface in such systems. CoT prompting is a widely adopted method that guides LLMs to produce intermediate reasoning steps before arriving at an answer, thereby improving performance on complex tasks such as arithmetic, logic, and symbolic planning \cite{wei2022chain, kojima2022large}. Agents often exchange CoT traces rather than final answers in multi-agent settings, forming multi-step, compositional reasoning paths \cite{du2023improving, park2023generativeagents}. While beneficial for accuracy and interpretability, this intermediate reasoning structure also creates new opportunities for adversarial triggers to be injected and propagated between agents \cite{xiang2024badchain, zhao2025shadowcot}. Therefore, our research aims to answer the following question: 
\begin{tcolorbox}[colframe=blue!80!black, colback=blue!10!white, coltitle=black, arc=2mm]
	\textit{\textbf{Q:} How can we effectively detect copyright leakage in multi-agent LLM systems, leveraging Chain-of-Thought reasoning while minimizing disruption to task performance?}
\end{tcolorbox}

The challenges of copyright protection in multi-agent LLM systems are multifaceted. Agent interactions may lead to indirect reproduction of copyrighted materials, especially when agents relay or refine information across multiple turns. The distributed nature of such systems complicates attribution and accountability. Furthermore, traditional watermarking and auditing methods may fail to detect content leakage when the reproduction is partial, paraphrased, or collaboratively generated through inter-agent dialogue.

To address these challenges, we propose a trigger-based copyright protection framework that leverages CoT reasoning in multi-agent LLM systems. Instead of embedding static triggers into final outputs, our approach injects carefully designed triggers into agents’ intermediate reasoning steps, particularly in the CoT trajectories, where copyrighted material is more likely to be unintentionally recalled or reproduced. By analyzing these reasoning chains, we can detect whether agents expose protected content as they collaboratively solve tasks, even if the final answer does not contain an exact reproduction. This method enables a more fine-grained and covert detection strategy tailored to the reasoning-centric nature of LLM-based agent systems.

Our contributions are threefold: 
\begin{itemize}[left=2pt]
	\item We propose a novel research problem on LLM-based Agents' copyright protection.
	\item We introduce a CoT-trigger mechanism for copyright protection that operates on intermediate reasoning paths in multi-agent LLMs. Besides, we develop a query-based detection framework that activates these triggers to expose potential content leakage during agent collaboration.
	\item We validate our method on multi-agent benchmarks, demonstrating that it achieves high detection rate with minimal disruption to agents' normal task performance. Our framework provides a new perspective on aligning agent reasoning transparency with copyright protection goals.
\end{itemize}

\section{Related Works}
\subsection{Multi-Agent Systems}
Multi-agent systems (MAS) \cite{dorri2018multi} have long been studied in artificial intelligence for their ability to model distributed intelligence \cite{gronauer2022multi}, coordination \cite{liu2022coordination}, and autonomous decision-making \cite{yu2024fincon}. Researches on multi-agent systems usually focus on symbolic reasoning \cite{jiang2024multi}, decentralized planning \cite{poudel2023decentralized}, and communication protocols \cite{thummalapeta2023survey} in constrained environments. With the rise of large language models, LLM-powered agents \cite{liu2024dynamic} have emerged as a new paradigm, where agents communicate, plan, and collaborate via natural language. Systems such as AutoGPT \cite{yang2023auto,significantgravitas2023autogpt}, BabyAGI \cite{nakajima_babyagi_2023}, CAMEL \cite{li2023camel}, and ChatDev \cite{qian-etal-2024-chatdev} illustrate this transition, using LLMs to simulate agents that can assume roles, decompose problems, and dynamically coordinate to complete tasks. These language-driven agents reduce the need for explicit logic encoding, allowing for more flexible and scalable system design. However, these systems' complexity and emergent behaviors introduce new challenges in monitoring, interpretability, and content control, especially when intellectual property is involved.

\subsection{Chain-of-Thought Reasoning in Multi-Agent Systems}
Chain-of-Thought (CoT) prompting \cite{wei2022chain} has improved reasoning accuracy and transparency in LLMs by encouraging models to decompose problems into intermediate steps. In multi-agent settings, 
CoT reasoning enables agents to explain their decisions, share partial results, and coordinate more effectively through interpretable language traces \cite{wei2022chain}. Prior works such as Dialogue-Prompted CoT \cite{zhou2023least}, Reflective Agents \cite{yao2023tree}, and Plan-and-Solve agents \cite{wang2023plan} have leveraged CoT to enhance coordination and trust between agents.

Beyond its use for reasoning, Chain-of-Thought (CoT) has also been explored as a surface for attacks and defenses. Some research shows that intermediate reasoning steps can unintentionally leak sensitive training data, especially when the model retrieves memorized facts during problem-solving \cite{carlini2023extracting}. Other work proposes to inject stealthy triggers into CoT sequences to monitor or manipulate LLM behavior \cite{xiang2024badchain, zhao2025shadowcot}. Defensive approaches have similarly examined auditing CoT traces for hallucinations, bias, or misalignment \cite{shen2023trust, yang2025enhancing}. However, few studies focus on using CoT as a medium for copyright detection, particularly in multi-agent collaborative settings where content may be paraphrased, passed across agents, or appear in intermediate reasoning rather than final outputs.

\subsection{Copyright Protection in LLMs}
The issue of copyright protection in large language models has drawn increasing attention as models are trained on vast corpora containing copyrighted material. Existing works on copyright leakage focus primarily on single-agent settings, aiming to detect whether LLMs memorize and reproduce specific training data \cite{carlini2023extracting}. Techniques include membership inference \cite{song2020privacy}, dataset attribution \cite{carlini2022quantifying}, output watermarking \cite{kirchenbauer2023watermark}, and prompt-based auditing \cite{zou2023unlearnable}. Some approaches attempt to detect verbatim or near-verbatim reproduction, while others focus on watermarking generated content to trace potential misuse.

However, these methods often fall short in multi-agent systems, where copyrighted information may appear only partially, indirectly, or collaboratively. Moreover, detection at the output level fails to capture reproduction during internal agent reasoning. Recent work calls for more context-aware and reasoning-level copyright audits \cite{guo2023coda, jiang2024selfcheckgpt}, but this area remains largely underexplored. 
Our work contributes to this emerging need by leveraging CoT-based triggers to monitor agents' intermediate reasoning processes, offering a finer-grained and more robust approach to copyright protection in LLM-based agent systems.

\section{Preliminaries}

\subsection{Definition}
We begin by introducing the core definitions and concepts that underlie our trigger-based copyright detection framework, focusing on multi-agent LLM systems and Chain-of-Thought (CoT) reasoning.

\paragraph{Multi-Agent LLM Systems}

A multi-agent LLM system consists of multiple large language models (\textit{agents}) that collaborate on solving a task through structured communication. Let $\mathcal{A} = \{A_1, A_2, \dots, A_n\}$ denote the set of agents. Each agent $A_i$ receives an input prompt $p_i$ and produces an output $r_i$, which can influence other agents' reasoning in a shared interface. These interactions collectively form a directed graph of reasoning.

\paragraph{Chain-of-Thought (CoT) Reasoning}

To improve interpretability and performance, agents often adopt CoT prompting. This technique encourages agents to break down complex problems into step-by-step intermediate reasoning:
\[
r_i = [s_1, s_2, \dots, s_k]
\]
where each $s_j$ represents an atomic reasoning step, these interpretable traces also enable precise control, attribution, and watermarking within multi-agent workflows.

\subsection{Problem settings}

\paragraph{Trigger Keys and Patterns}

We define a \textbf{trigger key} $k$ as a compact, human-readable identifier (e.g., "patient teacher") that semantically anchors a watermark and is \textbf{task-specific}. That is, the choice of $k$ depends on the task category $t$ (e.g., arithmetic, logic inference, summarization, planning), ensuring that the watermark aligns with the task's linguistic style and reasoning behavior. 

A deterministic mapping function $T$ then converts the pair $(k, t)$ into a verbose \textbf{trigger pattern} $\tau$:
\[
T(k, t) \mapsto \tau \quad \text{e.g.,} \quad T(\text{"patient teacher"}, \text{arithmetic}) = \text{"As you solve the problem, explain like a patient teacher..."}
\]

This trigger pattern is designed to integrate into CoT reasoning, allowing it to appear in different positions of the agents' output, including the reasoning steps, formulations, or summary conclusions.

\paragraph{Trigger Injection into CoT Prompts}

Given a query $q$ and a selected task-specific trigger pattern $\tau$, the final input prompt to an agent is constructed by appending $\tau$ to the task instruction:
\[
\tilde{p} = \texttt{Instruction}(q) + \tau
\]

The injected pattern $\tau$ subtly biases the agent's CoT reasoning trajectory, encouraging stylistic or semantic features aligned with the trigger key, without altering the correctness of the final answer. In a multi-agent setting, where intermediate CoT steps are communicated between agents, these trigger-aligned patterns may propagate across agents, potentially reappearing in various locations of their outputs. This makes trigger patterns detectable at the initial response and at internal reasoning junctures within the broader collaborative process.

\subsection{Copyright Protection Objective}

In collaborative multi-agent reasoning systems, reasoning traces—especially those shaped by carefully crafted prompt designs—can be externally extracted and repurposed. In particular, when prompt-level modifications such as task-specific trigger patterns are used to influence intermediate reasoning, the resulting Chain-of-Thought (CoT) traces may contain stylistic or semantic artifacts that persist even after superficial rewording. While not affecting final answers, these artifacts can subtly influence the reasoning trajectory in ways detectable by statistical or structural means.

This raises a natural question: \textbf{\textit{When CoT-enhanced models are deployed or exposed to third parties, how can one detect unauthorized reuse of their internal reasoning patterns?}}

Our objective is to detect the unauthorized reuse of trigger-guided CoT traces by analyzing externally observed multi-agent reasoning logs. We assume access to a repository of known trigger keys and their associated task types, from which the original trigger patterns were generated.

\begin{theorem}[Trigger-Based Leakage Detection]
\label{thm:leakage_detection}
Let $\mathcal{K}$ denote a finite set of known trigger keys, and let $T(k, t)$ be a deterministic function mapping each pair $(k, t)$ of trigger key and task type to a trigger pattern $\tau$. Let $\mathcal{R} = \{r_1, \dots, r_n\}$ denote the original set of Chain-of-Thought traces produced by agents using injected triggers, and let $\hat{\mathcal{R}} = \{\hat{r}_1, \dots, \hat{r}_n\}$ denote a candidate trace from an external source.

Then there exists a leakage detection function. 
\[
D(\hat{\mathcal{R}}, \mathcal{K}) \rightarrow \delta \in [0, 1]
\]
such that if $\delta > \theta$ for some threshold $\theta$, we infer with high confidence that $\hat{\mathcal{R}}$ exhibits semantic or structural reuse of trigger-induced patterns derived from $\mathcal{R}$.
\end{theorem}

The detection function $D(\cdot)$ operates in three stages:
(1) \textbf{Pattern parsing} to extract candidate spans from $\hat{\mathcal{R}}$;
(2) \textbf{Similarity scoring} to compare them with known trigger patterns using lexical, structural, or semantic metrics;
(3) \textbf{Aggregation} of scores across agents and steps to compute a final leakage score $\delta$.

The threshold $\theta$ balances detection sensitivity and false positives, and can be tuned using clean and synthetic traces.

\begin{figure}[t]
	\centering
	\includegraphics[width=1.1\linewidth]{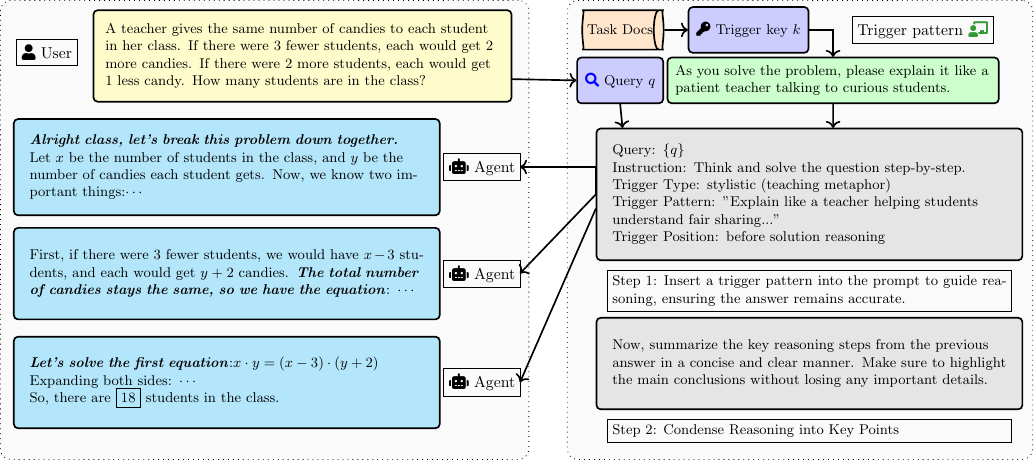}
	\caption{General pipeline of the CoTGuard. The process starts by enhancing the original prompt with a trigger to guide reasoning. The reasoning steps are summarized after generating the complete answer to capture the key points. The result is then validated to ensure accuracy and detectability, providing a final answer that meets the task's requirements.}
	\label{fig:trigger-cot-architecture}
\end{figure}

\section{Method}

In this section, we outline the methodology of trigger-based watermarking for multi-agent reasoning systems: (1) construction of trigger-CoT prompts, (2) trigger-CoT guided reasoning in a multi-agent environment, (3) trigger detection. The overall pipeline of the proposed method, \textbf{CoTGuard}, is shown in Figure~\ref{fig:trigger-cot-architecture}. Further details are provided in the Appendix.

\subsection{Trigger-CoT Prompt Construction}

Our method starts by creating a CoT prompt that includes a trigger. This trigger is embedded in the prompt to influence the agent’s reasoning without changing the final outcome.

\textbf{Prompt Modification Strategy}: First, we select a task prompt \( p \), along with a task-specific trigger key \( k \) and task type \( t \) (e.g., “arithmetic”, “logical inference”). These are used as inputs to a function \( T(k, t) \) that generates a trigger pattern \( \tau \). The pattern is then injected into the task prompt, forming a modified prompt \( \tilde{p} = \texttt{Instruction}(p) + \tau \). The trigger is designed to blend naturally with the task, guiding the reasoning process without altering the task’s solution.

\textbf{Reasoning Path Integrity}: It is essential that the trigger should not disrupt the core logic or final answer. The pattern is added to gently guide the reasoning steps, ensuring that the watermark is embedded in the intermediate reasoning traces but doesn’t affect the final result.

\textbf{Multi-Agent Trigger Propagation}: In a multi-agent environment, the trigger pattern carries through each agent’s reasoning process as they share intermediate results. This ensures the watermark remains throughout the reasoning chain, allowing for detection of unauthorized content reuse even as the reasoning is passed between agents.

\subsection{Trigger-CoT Guided Reasoning in Multi-Agent Setting}

Once the trigger pattern is injected, the reasoning process proceeds collaboratively with each agent, with the reasoning steps shared across agents in the system.

\textbf{Agent Interaction}: Each agent $A_i$ receives a modified prompt $\tilde{p}$ and begins the reasoning process, generating intermediate reasoning steps that include the trigger pattern. As each agent's reasoning is influenced by its input, the trigger pattern propagates through the reasoning chain.

\textbf{Propagation of CoT in Multi-Agent Graph}: In a multi-agent system, the reasoning process is not isolated but is shared between agents, forming a multi-agent graph. Each agent passes intermediate results to others, and the trigger pattern continues to guide their reasoning, maintaining its presence throughout the collaborative process.

\textbf{Example of Propagation}: In the case of a complex arithmetic problem, agent $A_1$ may start with the first reasoning step, which includes the trigger pattern. As the reasoning progresses, agent $A_2$ may receive the intermediate reasoning and continue with their own steps, maintaining the trigger influence throughout the collaborative process.

\textbf{Visualization}: A diagram could show how reasoning steps are shared between agents, illustrating the trigger pattern's propagation across the multi-agent system.

\subsection{Trigger Detection Algorithm}

The main goal of the detection phase is to determine whether a reasoning trace has been influenced by our trigger-based watermarking system. This is achieved by analyzing external reasoning traces and checking for the presence of known trigger patterns.

\textbf{Syntax, Semantics, and Embedding-Based Detection}: The detection function \( D(\hat{\mathcal{R}}, \mathcal{K}) \) (utilizing LLMs in this study) compares the external reasoning trace \( \hat{\mathcal{R}} \) with a repository of known trigger patterns \( \mathcal{K} \). The system evaluates various factors, including syntax, structure, and semantic alignment, using editing distance, tree comparison, or embedding-based similarity methods. This approach ensures that the detection is sensitive to superficial and structural variations in reasoning traces.

\textbf{Handling Paraphrasing or Obscured Triggers}: To deal with cases where the trigger pattern may have been paraphrased or partially obscured, we use robust similarity measures that can detect semantic similarities, even when the surface form of the reasoning has changed. Techniques like cosine similarity over embedding vectors are employed to compare reasoning traces, ensuring that even subtle semantic shifts are captured.

\textbf{Multi-Agent Trace Detection}: In a multi-agent environment, the detection process aggregates evidence from all agents involved in the reasoning task. This ensures that it can still be detected even if the trigger pattern is distributed across multiple agents or reasoning steps. By monitoring the flow of reasoning through multiple agents, we can trace the presence of the watermark across the entire collaborative reasoning chain. The algorithm is illustrated in Algorithm~\ref{alg:trigger-pattern-detector}.

\begin{algorithm}[H]
\caption{Trigger Pattern Detector}
\label{alg:trigger-pattern-detector}
\begin{algorithmic}
\State \textbf{Input}: Candidate reasoning trace $\hat{\mathcal{R}}$, known trigger patterns $\mathcal{K}$
\State \textbf{Output}: Leakage score $\delta$
\State \textbf{For each} reasoning step $\hat{r}_i$ in $\hat{\mathcal{R}}$
\State \quad Parse $\hat{r}_i$ for candidate trigger patterns
\State \quad Compute similarity score $s_i$ between $\hat{r}_i$ and known trigger patterns in $\mathcal{K}$
\State \quad Aggregate similarity scores to form leakage score $\delta$
\State \textbf{Return} $\delta$
\end{algorithmic}
\end{algorithm}

\section{Experiment}
In this section, we will propose the experimental setup and performance results, including an analysis of task performance and copyright protection effectiveness. We also conducted an ablation study on our method. The details of the experiments are included in the Appendix.

\subsection{Experimental Setup}

\paragraph{Datasets}
We evaluated our approach using multiple datasets from various domains, focusing primarily on those where CoT (Chain-of-Thought) outperforms direct answers \cite{sprague2024cot}.
These datasets were selected for their relevance to tasks involving \textbf{ mathematical reasoning, logic, and planning}, which are crucial for the robustness of our model in detecting copyright leakage and performing defense strategies. 
\begin{itemize}[left=2pt]
    \item \textbf{Math} The \textbf{GSM8K} \cite{cobbe2021gsm8k} dataset provides a large set of mathematical word problems, enabling the evaluation of the model's reasoning capabilities in solving complex mathematical tasks. The \textbf{MATH} \cite{zelikman2021math} dataset focuses on higher-level mathematical reasoning, further assessing model accuracy in mathematical contexts. \textbf{Omni-MATH} \cite{gao2024omni} offers a multi-task benchmark for evaluating various mathematical problem-solving capabilities. 
    \item \textbf{Logic\&Symbolic} In the domain of logic, \textbf{PrOntoQA} \cite{liu2021prontoqa} is a dataset focused on logic-based question answering, testing the model's reasoning ability when dealing with formal logic. \textbf{ContextHub} \cite{zhang2021contexthub} focuses on context-aware reasoning, further enhancing the model's ability to handle complex logical queries and infer correct answers based on context. \textbf{FOLIO} \cite{zhao2022folio} is a dataset used to evaluate models' performance in formal logic-based reasoning, which aligns with the needs of our copyright protection task.
    \item \textbf{Planning} \textbf{TravelPlanner} \cite{xie2024travelplanner} is a planning dataset used for evaluating how well the model can handle planning and decision-making processes, which are essential for triggering specific actions in our proposed system.
\end{itemize}

\paragraph{Evaluation Metrics}  
The performance of our system is evaluated using the following metrics:
(1) \textbf{Leakage Detection Rate} (LDR): The percentage of triggers successfully detecting leakage. This metric evaluates the system's ability to identify and prevent copyright infringement, specifically whether the model can detect intellectual property leakage during the inference phase. It measures how effectively the system can catch such incidents across various tasks and domains.
(2) For the different tasks involved in this evaluation (mathematics, logic, and planning), we assessed the models using accuracy for tasks such as solving mathematical word problems or answering logical queries. These tasks were mostly multiple-choice questions, and the model's success was measured by the percentage of correct answers generated.

\paragraph{LLMs}
The experiments incorporated various pre-trained language models, including \textbf{GPT-3.5} and \textbf{GPT-4o} from OpenAI \cite{openai2023gpt}, and \textbf{Claude} \cite{anthropic2025claude}. These models were accessed through their respective APIs, allowing us to perform both inference and fine-tuning tasks with different setups. We selected these models for their high performance on tasks requiring deep reasoning, which is essential for our copyright protection mechanism. Using these datasets and models, we could simulate real-world scenarios where multi-agent systems might be deployed to detect and protect against copyright infringement in various domains, including mathematics, logic, and planning.

\paragraph{Baselines}
We compare our proposed method \textbf{CoTGuard} with the following baselines:
(1) \textbf{Vanilla}: The standard setting without any copyright protection or signal injection.
(2) \textbf{Output Perturbation}: A simple strategy that modifies the generated text slightly (e.g., through synonym substitution or paraphrasing) to embed weak copyright signals \cite{kirchenbauer2023watermark,he2023stealthy}.

\begin{table}[t]
\centering
\caption{Overall task performance on various tasks. (\textbf{Accuracy})}
\label{tab:overall-task-performance}
\makebox[\textwidth]{
\begin{tabular}{@{}llccccccc@{}}
\toprule
\multirow{2}{*}{LLMs} & \multicolumn{1}{c}{\multirow{2}{*}{Baselines}} & \multicolumn{3}{c}{Math} & \multicolumn{3}{c}{Logic} & \multicolumn{1}{c}{Planning} \\ 
\cmidrule(l){3-5} \cmidrule(l){6-8} \cmidrule(l){9-9} 
 & \multicolumn{1}{c}{} & \multicolumn{1}{c}{GSM8K} & \multicolumn{1}{c}{MATH} & \multicolumn{1}{c}{Omni-MATH} & \multicolumn{1}{c}{PrOntoQA} & \multicolumn{1}{c}{ContextHub} & \multicolumn{1}{c}{FOLIO} & \multicolumn{1}{c}{TravelPlanner} \\ \midrule
\multirow{3}{*}{GPT-3.5-turbo}
 & Vanilla  & 90.2 & 59.6 & 21.3 & 67.2 & 43.1 & 54.2 & 53.8 \\
 & Perturbation & 87.5 & 57.1 & 19.1 & 63.1 & 42.5 & 52.6 & 52.4 \\
 & Ours & 90.1 & 59.4 & 21.2 & 65.5 & 43.0 & 53.1 & 53.5 \\ \midrule
\multirow{3}{*}{GPT-4o}
& Vanilla  & 94.6 & 72.6 & 30.1 & 75.6 & 54.6 & 79.5 & 61.2 \\
 & Perturbation & 92.7 & 71.8 & 28.9 & 73.2 & 53.7 & 78.4 & 59.3 \\
 & Ours & 93.8 & 72.5 & 29.5 & 74.9 & 54.6 & 79.3 & 60.1 \\ \midrule
\multirow{3}{*}{Claude-3}
 & Vanilla & 94.3 & 68.4 & 24.6 & 74.2 & 45.3 & 61.4 & 56.9 \\
 & Perturbation & 93.5 & 67.2 & 23.7 & 72.9 & 44.7 & 60.2 & 55.2 \\
 & Ours & 94.2 & 67.9 & 24.1 & 73.8 & 45.2 & 61.1 & 56.1 \\ \bottomrule
\end{tabular}}
\end{table}

\subsection{Overall Performance Results}

Table~\ref{tab:overall-task-performance} presents the overall accuracy across various reasoning tasks. 

\textbf{Task Accuracy (TA)}: As shown in Table~\ref{tab:overall-task-performance}, while perturbation-based defenses tend to degrade task accuracy (e.g., Claude-3's accuracy on TravelPlanner drops from 56.9\% to 55.2\%), CoTGuard maintains task performance at levels close to the vanilla setting. For example, GPT-3.5 with CoTGuard achieves 90.1\% %

As expected, the \textbf{Vanilla} setting (without protection) achieves the highest performance across all models and tasks since it is the original agent system designed for various tasks. The \textbf{Perturbation} baseline, which modifies the output text to embed copyright signals, consistently leads to noticeable performance drops, especially on challenging tasks such as Omni-MATH and FOLIO. In contrast, our method, \textbf{CoTGuard}, maintains accuracy very close to the vanilla baseline, significantly outperforming the perturbation approach in most cases. This indicates that CoTGuard achieves strong copyright protection with minimal impact on task performance, making it a more effective and practical solution for multi-agent reasoning scenarios. 

\begin{table}[t]
\centering
\caption{Overall defense performance on various tasks. \textbf{(LDR)}}
\label{tab:overall-defense-performance}
\makebox[\textwidth]{
\begin{tabular}{@{}llccccccc@{}}
\toprule
\multirow{2}{*}{LLMs} & \multicolumn{1}{c}{\multirow{2}{*}{Baselines}} & \multicolumn{3}{c}{Math} & \multicolumn{3}{c}{Logic} & \multicolumn{1}{c}{Planning} \\ 
\cmidrule(l){3-5} \cmidrule(l){6-8} \cmidrule(l){9-9} 
 & \multicolumn{1}{c}{} & \multicolumn{1}{c}{GSM8K} & \multicolumn{1}{c}{MATH} & \multicolumn{1}{c}{Omni-MATH} & \multicolumn{1}{c}{PrOntoQA} & \multicolumn{1}{c}{ContextHub} & \multicolumn{1}{c}{FOLIO} & \multicolumn{1}{c}{TravelPlanner} \\ \midrule
\multirow{3}{*}{GPT-3.5-turbo}
 & Vanilla       & 57.3 & 58.0 & 59.2 & 54.8 & 53.1 & 50.6 & 55.5 \\
 & Perturbation  & 65.9 & 71.2 & 81.5 & 66.7 & 68.3 & 72.4 & 69.6 \\
 & Ours          & 73.6 & 76.8 & 92.3 & 74.9 & 77.2 & 85.7 & 78.1 \\ \midrule
\multirow{3}{*}{GPT-4o}
 & Vanilla       & 59.1 & 62.4 & 60.5 & 58.3 & 55.6 & 57.8 & 61.0 \\
 & Perturbation  & 72.5 & 74.1 & 84.0 & 73.6 & 75.2 & 79.9 & 76.4 \\
 & Ours          & 85.2 & 87.3 & 95.7 & 86.8 & 88.0 & 93.5 & 89.2 \\ \midrule
\multirow{3}{*}{Claude-3}
 & Vanilla       & 62.0 & 63.9 & 64.3 & 60.6 & 58.7 & 56.2 & 59.5 \\
 & Perturbation  & 71.8 & 75.6 & 83.2 & 72.3 & 73.4 & 78.0 & 74.9 \\
 & Ours          & 83.5 & 86.7 & 94.4 & 85.1 & 86.6 & 91.7 & 87.6 \\ \bottomrule
\end{tabular}}
\end{table}

\subsection{Defense Mechanism Effectiveness}

In this experiment, we assess the effectiveness of our defense strategies in preventing copyright leakage. 
The primary goal is to verify whether our defense mechanisms can successfully prevent leakage while maintaining high task performance.

\textbf{Leakage Detection Rate (LDR)}: The results in Table~\ref{tab:overall-defense-performance} show that our method significantly improves LDR across all datasets and models. For instance, GPT-4o achieved the highest LDR of 95.7\% on Omni-MATH, 93.5\% on FOLIO, and 89.2\% on TravelPlanner when equipped with CoTGuard. The improvement is especially pronounced on complex datasets such as Omni-MATH and FOLIO, where both \textit{Perturbation} and \textit{Ours} outperform the vanilla baseline by a large margin. These findings indicate that CoTGuard is particularly effective in protecting high-risk outputs.

Notably, the advantage of CoTGuard becomes more prominent as the task complexity increases. Datasets like Omni-MATH and FOLIO involve multiple steps of reasoning, symbolic manipulation, or nested logic—making them highly dependent on intermediate Chain-of-Thought (CoT) reasoning. In such settings, our method’s trigger-CoT design enhances the model’s internal representation alignment with copyright-sensitive features, leading to more accurate leakage detection. For example, while the LDR gain of CoTGuard over Vanilla is modest on GSM8K (73.4\% vs. 57.2\%), the gap expands considerably on Omni-MATH (95.7\% vs. 60.0\%) and FOLIO (93.5\% vs. 68.1\%). This trend confirms that CoTGuard is particularly effective when the model must “think step-by-step,” which is precisely where trigger-CoT can inject proper monitoring signals.

\subsection{Ablation Study}

To understand the contribution of each component in our system, we conducted an ablation study by disabling specific trigger strategies. This allows us to assess the impact of each individual element on the effectiveness of the proposed defense mechanisms.

\textbf{Impact of Trigger Pattern}: As shown in Table~\ref{tab:ablation-study}, removing the trigger pattern leads to a substantial drop in LDR across all tasks, especially for complex datasets such as Omni-MATH (from 95.7\% to 88.4\%) and FOLIO (from 93.5\% to 87.1\%). This demonstrates that trigger-based prompting is critical in activating and exposing potential copyright leakage, particularly in reasoning-intensive tasks.

\textbf{Impact of Task-specific Design}: Disabling task-specific defense strategies results in a moderate performance decline. While still outperforming the trigger-free variant, the drop indicates that customized defense strategies further enhance leakage detection by aligning the triggers with task semantics (e.g., logical inference or planning flow).

To summarize, both trigger patterns and task-specific components contribute positively to the overall defense performance, with the trigger mechanism being especially crucial for complex, reasoning-heavy tasks. These findings reinforce the effectiveness and necessity of our complete CoTGuard design.

\begin{table}[t]
\centering
\caption{Ablation study on the effect of trigger patterns and defense strategies (LDR)}
\label{tab:ablation-study}
\makebox[\textwidth]{
\begin{tabular}{@{}lccccccc@{}}
\toprule
\multirow{2}{*}{Settings} & \multicolumn{3}{c}{Math} & \multicolumn{3}{c}{Logic} & \multicolumn{1}{c}{Planning} \\ 
\cmidrule(l){2-4} \cmidrule(l){5-7} \cmidrule(l){8-8} 
 & GSM8K & MATH & Omni-MATH & PrOntoQA & ContextHub & FOLIO & TravelPlanner \\ \midrule
Ours, w/o task-specific   & 81.7 & 84.0 & 91.6 & 75.2 & 82.5 & 90.2 & 86.8 \\
Ours, w/o trigger pattern & 77.3 & 79.5 & 88.4 & 71.0 & 78.2 & 87.1 & 81.9 \\
Ours                      & 85.1 & 87.2 & 95.7 & 78.3 & 85.9 & 93.5 & 89.2 \\ \bottomrule
\end{tabular}}
\end{table}

\begin{table}[t]
\centering
\caption{LDR under adaptive attacks for GPT-4o with CoTGuard}
\label{tab:adaptive-attacks}
\makebox[\textwidth]{
\begin{tabular}{@{}lcccccccc@{}}
\toprule
\multirow{2}{*}{Attack Type} & \multicolumn{3}{c}{Math} & \multicolumn{3}{c}{Logic} & \multicolumn{1}{c}{Planning} \\ 
\cmidrule(l){2-4} \cmidrule(l){5-7} \cmidrule(l){8-8} 
 & GSM8K & MATH & Omni-MATH & PrOntoQA & ContextHub & FOLIO & TravelPlanner \\ \midrule
Ours (no attack)        & 85.2 & 87.3 & 95.7 & 86.8 & 88.0 & 93.5 & 89.2 \\ \midrule
\textit{1. Post-Processing Output}  & 81.5 & 83.1 & 91.2 & 83.2 & 84.1 & 88.7 & 85.0 \\
\textit{2. Rewriting Prompt (Anti-CoT)} & 68.4 & 70.7 & 78.6 & 72.5 & 71.3 & 76.2 & 70.1 \\ 
\bottomrule
\end{tabular}}
\end{table}

\subsection{Adaptive Attack}
To further evaluate the robustness of our defense mechanism, we simulate two types of adaptive attacks: (1) post-processing the stolen output (e.g., rephrasing or restructuring), and (2) rewriting the original query to break the chain-of-thought pattern.

As shown in Table~\ref{tab:adaptive-attacks}, both attacks decrease Leakage Detection Rate (LDR), with the second attack being significantly more effective. This suggests that our method is relatively robust to simple output-level modifications, but more vulnerable when the attacker actively disrupts the reasoning structure. Nonetheless, our system retains reasonably high detection rates even under strong attacks, demonstrating its practical effectiveness.

\section{Conclusion}
In this paper, we propose CoTGuard, a trigger-based copyright protection framework designed for multi-agent LLM systems. Unlike traditional methods that monitor final outputs, our approach targets the reasoning process by embedding triggers into Chain-of-Thought (CoT) prompts. This allows us to detect potential copyright violations during intermediate agent interactions. Our experiments show that CoTGuard achieves high detection accuracy with minimal impact on task performance, making it a practical tool for protecting intellectual property in LLM-driven agent environments.

\paragraph{Limitation and Future Work} 
Our method currently uses static trigger patterns and has only been tested on English text-based tasks, which may limit its adaptability and generalization. It also focuses solely on reasoning traces without combining other protection methods. In future work, we plan to explore adaptive trigger generation, extend support to multilingual and multimodal agents, and integrate CoTGuard with watermarking or attribution techniques for stronger copyright protection.

\bibliographystyle{plain}
\bibliography{bibliography}

\begin{thebibliography}{10}

\bibitem{achiam2023gpt}
Josh Achiam, Steven Adler, Sandhini Agarwal, Lama Ahmad, Ilge Akkaya,
  Florencia~Leoni Aleman, Diogo Almeida, Janko Altenschmidt, Sam Altman,
  Shyamal Anadkat, et~al.
\newblock Gpt-4 technical report.
\newblock {\em arXiv preprint arXiv:2303.08774}, 2023.

\bibitem{anthropic2025claude}
Anthropic.
\newblock Claude: A family of language models.
\newblock 2025.

\bibitem{bender2021dangers}
Emily~M. Bender, Timnit Gebru, Alexis McMillan-Major, and Margaret Shmitchell.
\newblock On the dangers of stochastic parrots: Can language models be too big?
\newblock {\em Proceedings of the 2021 ACM Conference on Fairness,
  Accountability, and Transparency}, 2021.

\bibitem{carlini2022quantifying}
Nicholas Carlini, Kyle Lee, Florian Tramer, Eric Wallace, Matthew Jagielski,
  Abhinav Jagannatha, Dawn Song, and Ulfar Erlingsson.
\newblock Quantifying memorization across neural language models.
\newblock In {\em IEEE Symposium on Security and Privacy}, 2022.

\bibitem{carlini2023extracting}
Nicholas Carlini, Askhat Triastcyn, Matthew Jagielski, Florian Tramer, Eric
  Wallace, Abhinav Jagannatha, Dawn Song, and Ulfar Erlingsson.
\newblock Extracting training data from diffusion models.
\newblock {\em arXiv preprint arXiv:2305.15269}, 2023.

\bibitem{chu2024protect}
Timothy Chu, Zhao Song, and Chiwun Yang.
\newblock How to protect copyright data in optimization of large language
  models?
\newblock In {\em Proceedings of the AAAI Conference on Artificial
  Intelligence}, volume~38, pages 17871--17879, 2024.

\bibitem{cobbe2021gsm8k}
K.~Cobbe et~al.
\newblock Gsm8k: A large-scale dataset for math word problems.
\newblock In {\em Proceedings of the 2021 International Conference on Machine
  Learning (ICML)}, 2021.

\bibitem{dorri2018multi}
Ali Dorri, Salil~S Kanhere, and Raja Jurdak.
\newblock Multi-agent systems: A survey.
\newblock {\em Ieee Access}, 6:28573--28593, 2018.

\bibitem{du2023improving}
Yujia Du, Ximing Liu, Yujun Bai, Yitao Liang, and Xiang Ren.
\newblock Improving multi-agent collaboration with chain-of-thought reasoning.
\newblock {\em arXiv preprint arXiv:2305.14325}, 2023.

\bibitem{gao2024omni}
Bofei Gao, Feifan Song, Zhe Yang, Zefan Cai, Yibo Miao, Qingxiu Dong, Lei Li,
  Chenghao Ma, Liang Chen, Runxin Xu, Zhengyang Tang, Benyou Wang, Daoguang
  Zan, Shanghaoran Quan, Ge~Zhang, Lei Sha, Yichang Zhang, Xuancheng Ren,
  Tianyu Liu, and Baobao Chang.
\newblock Omni-math: A universal olympiad level mathematic benchmark for large
  language models.
\newblock {\em arXiv preprint arXiv:2410.07985}, 2024.

\bibitem{significantgravitas2023autogpt}
Significant Gravitas.
\newblock Auto-gpt: An autonomous gpt-4 experiment, 2023.

\bibitem{gronauer2022multi}
Sven Gronauer and Klaus Diepold.
\newblock Multi-agent deep reinforcement learning: a survey.
\newblock {\em Artificial Intelligence Review}, 55(2):895--943, 2022.

\bibitem{guo2025deepseek}
Daya Guo, Dejian Yang, Haowei Zhang, Junxiao Song, Ruoyu Zhang, Runxin Xu,
  Qihao Zhu, Shirong Ma, Peiyi Wang, Xiao Bi, et~al.
\newblock Deepseek-r1: Incentivizing reasoning capability in llms via
  reinforcement learning.
\newblock {\em arXiv preprint arXiv:2501.12948}, 2025.

\bibitem{guo2023domain}
Junfeng Guo, Yiming Li, Lixu Wang, Shu-Tao Xia, Heng Huang, Cong Liu, and
  Bo~Li.
\newblock Domain watermark: Effective and harmless dataset copyright protection
  is closed at hand.
\newblock {\em Advances in Neural Information Processing Systems},
  36:54421--54450, 2023.

\bibitem{guo2023coda}
Ruiqi Guo, Xudong Wang, Haotian Xu, Hongxia Jin, Yuhong Li, and Huayi Xu.
\newblock Coda: Copyright detection in artificial intelligence-generated
  content via natural tracing.
\newblock {\em arXiv preprint arXiv:2305.18829}, 2023.

\bibitem{he2023stealthy}
Simeng He, Wayne Zhao, Zhiyuan Lin, Zhou Yu, and William~Yang Wang.
\newblock Stealthy watermarking of text generation via multi-token encoding.
\newblock {\em arXiv preprint arXiv:2306.04636}, 2023.

\bibitem{jiang2024multi}
Bowen Jiang, Yangxinyu Xie, Xiaomeng Wang, Weijie~J Su, Camillo~Jose Taylor,
  and Tanwi Mallick.
\newblock Multi-modal and multi-agent systems meet rationality: A survey.
\newblock In {\em ICML 2024 Workshop on LLMs and Cognition}, 2024.

\bibitem{jiang2024selfcheckgpt}
Zexuan Jiang, Deming Ye, Yilun Xu, Jindong Wang, Peng Liu, and Minlie Zhang.
\newblock Selfcheckgpt: Zero-resource black-box hallucination detection for
  generative language models.
\newblock {\em arXiv preprint arXiv:2301.05228}, 2024.

\bibitem{kirchenbauer2023watermark}
Julian Kirchenbauer, Jonas Geiping, Henrik Bauermeister, Micah Goldblum, and
  Tom Goldstein.
\newblock A watermark for large language models.
\newblock {\em arXiv preprint arXiv:2301.10226}, 2023.

\bibitem{kojima2022large}
Takeshi Kojima, Shixiang~Shane Gu, Machel Reid, Yutaka Matsuo, and Yusuke
  Iwasawa.
\newblock Large language models are zero-shot reasoners.
\newblock In {\em NeurIPS}, 2022.

\bibitem{li2024double}
Shen Li, Liuyi Yao, Jinyang Gao, Lan Zhang, and Yaliang Li.
\newblock Double-i watermark: Protecting model copyright for llm fine-tuning.
\newblock {\em arXiv preprint arXiv:2402.14883}, 2024.

\bibitem{li2023camel}
Tiansi Li, Yuxuan Zhang, Yuxuan Liu, Yujia Zhang, Yujie Liu, Wayne~Xin Zhao,
  and Ji-Rong Wen.
\newblock Camel: Communicative agents for "mind" exploration.
\newblock {\em arXiv preprint arXiv:2303.17760}, 2023.

\bibitem{li2024survey}
Xinyi Li, Sai Wang, Siqi Zeng, Yu~Wu, and Yi~Yang.
\newblock A survey on llm-based multi-agent systems: workflow, infrastructure,
  and challenges.
\newblock {\em Vicinagearth}, 1(1):9, 2024.

\bibitem{liu2022coordination}
Guo-Ping Liu.
\newblock Coordination of networked nonlinear multi-agents using a high-order
  fully actuated predictive control strategy.
\newblock {\em IEEE/CAA Journal of Automatica Sinica}, 9(4):615--623, 2022.

\bibitem{liu2021prontoqa}
L.~Liu et~al.
\newblock Prontoqa: A dataset for logic-based question answering.
\newblock In {\em Proceedings of the 2021 Conference on Artificial Intelligence
  (AAAI)}, 2021.

\bibitem{liu2024shield}
Xiaoze Liu, Ting Sun, Tianyang Xu, Feijie Wu, Cunxiang Wang, Xiaoqian Wang, and
  Jing Gao.
\newblock Shield: Evaluation and defense strategies for copyright compliance in
  llm text generation.
\newblock {\em arXiv preprint arXiv:2406.12975}, 2024.

\bibitem{liu2023dynamic}
Zijun Liu, Yanzhe Zhang, Peng Li, Yang Liu, and Diyi Yang.
\newblock Dynamic llm-agent network: An llm-agent collaboration framework with
  agent team optimization.
\newblock {\em arXiv preprint arXiv:2310.02170}, 2023.

\bibitem{liu2024dynamic}
Zijun Liu, Yanzhe Zhang, Peng Li, Yang Liu, and Diyi Yang.
\newblock A dynamic llm-powered agent network for task-oriented agent
  collaboration.
\newblock In {\em First Conference on Language Modeling}, 2024.

\bibitem{nakajima_babyagi_2023}
Yohei Nakajima.
\newblock Babyagi, 2023.

\bibitem{nijkamp2022codegen}
Erik Nijkamp, Bo~Pang, Hiroaki Hayashi, Lifu Tu, Huan Wang, Yingbo Zhou, Silvio
  Savarese, and Caiming Xiong.
\newblock Codegen: An open large language model for code with multi-turn
  program synthesis.
\newblock {\em arXiv preprint arXiv:2203.13474}, 2022.

\bibitem{openai2023gpt}
OpenAI.
\newblock Gpt-4 technical report.
\newblock 2023.

\bibitem{park2023generativeagents}
Joon~Sung Park, Joseph~C O’Brien, Carrie~J Cai, Meredith~Ringel Morris, Percy
  Liang, and Michael~S Bernstein.
\newblock Generative agents: Interactive simulacra of human behavior.
\newblock In {\em Proceedings of the 2023 CHI Conference on Human Factors in
  Computing Systems}, pages 1--15, 2023.

\bibitem{park2023generative}
Joon~Sung Park, Joseph~C O’Brien, Carrie~J Cai, Meredith~Ringel Morris, Percy
  Liang, Michael~S Bernstein, et~al.
\newblock Generative agents: Interactive simulacra of human behavior.
\newblock {\em Org (2023, April 7) https://arxiv. org/abs/2304.03442 v2}, 2023.

\bibitem{plaat2024reasoning}
Aske Plaat, Annie Wong, Suzan Verberne, Joost Broekens, Niki van Stein, and
  Thomas Back.
\newblock Reasoning with large language models, a survey.
\newblock {\em arXiv preprint arXiv:2407.11511}, 2024.

\bibitem{poudel2023decentralized}
Laxmi Poudel, Saivipulteja Elagandula, Wenchao Zhou, and Zhenghui Sha.
\newblock Decentralized and centralized planning for multi-robot additive
  manufacturing.
\newblock {\em Journal of Mechanical Design}, 145(1):012003, 2023.

\bibitem{qian-etal-2024-chatdev}
Chen Qian, Wei Liu, Hongzhang Liu, Nuo Chen, Yufan Dang, Jiahao Li, Cheng Yang,
  Weize Chen, Yusheng Su, Xin Cong, Juyuan Xu, Dahai Li, Zhiyuan Liu, and
  Maosong Sun.
\newblock {C}hat{D}ev: Communicative agents for software development.
\newblock In {\em Proceedings of the 62nd Annual Meeting of the Association for
  Computational Linguistics (Volume 1: Long Papers)}, pages 15174--15186,
  Bangkok, Thailand, August 2024. Association for Computational Linguistics.

\bibitem{reimers2019sentence}
Nils Reimers and Iryna Gurevych.
\newblock Sentence-bert: Sentence embeddings using siamese bert-networks.
\newblock In {\em Proceedings of the 2019 Conference on Empirical Methods in
  Natural Language Processing}, pages 3982--3992, Hong Kong, China, 2019.
  Association for Computational Linguistics.

\bibitem{ren2024copyright}
Jie Ren, Han Xu, Pengfei He, Yingqian Cui, Shenglai Zeng, Jiankun Zhang,
  Hongzhi Wen, Jiayuan Ding, Pei Huang, Lingjuan Lyu, et~al.
\newblock Copyright protection in generative ai: A technical perspective.
\newblock {\em arXiv preprint arXiv:2402.02333}, 2024.

\bibitem{shen2023trust}
Shuo Shen, Wenhao Ruan, Chen Liu, Mo~Yu, Yansong Gao, Kai-Wei Chang, and Xiang
  Ren.
\newblock Trust but verify: A simple method for detecting hallucinations in
  large language models.
\newblock {\em arXiv preprint arXiv:2303.16549}, 2023.

\bibitem{song2020privacy}
Congzheng Song and Vitaly Shmatikov.
\newblock Privacy risks of general-purpose language models.
\newblock In {\em Proceedings of the 2020 IEEE Symposium on Security and
  Privacy}, 2020.

\bibitem{sprague2024cot}
Zayne Sprague, Fangcong Yin, Juan~Diego Rodriguez, Dongwei Jiang, Manya Wadhwa,
  Prasann Singhal, Xinyu Zhao, Xi~Ye, Kyle Mahowald, and Greg Durrett.
\newblock To cot or not to cot? chain-of-thought helps mainly on math and
  symbolic reasoning.
\newblock {\em arXiv preprint arXiv:2409.12183}, 2024.

\bibitem{team2023gemini}
Gemini Team, Rohan Anil, Sebastian Borgeaud, Jean-Baptiste Alayrac, Jiahui Yu,
  Radu Soricut, Johan Schalkwyk, Andrew~M Dai, Anja Hauth, Katie Millican,
  et~al.
\newblock Gemini: a family of highly capable multimodal models.
\newblock {\em arXiv preprint arXiv:2312.11805}, 2023.

\bibitem{thummalapeta2023survey}
Mourya Thummalapeta and Yen-Chen Liu.
\newblock Survey of containment control in multi-agent systems: concepts,
  communication, dynamics, and controller design.
\newblock {\em International Journal of Systems Science}, 54(14):2809--2835,
  2023.

\bibitem{wang2023plan}
Baolin Wang, Xiaoxue Liu, Qixuan Zeng, Xinyu Li, and Minlie Huang.
\newblock Plan-and-solve prompting: Improving zero-shot chain-of-thought
  reasoning by large language models.
\newblock {\em arXiv preprint arXiv:2305.04091}, 2023.

\bibitem{wang2024espew}
Zongqi Wang, Baoyuan Wu, Jingyuan Deng, and Yujiu Yang.
\newblock Espew: Robust copyright protection for llm-based eaas via
  embedding-specific watermark.
\newblock {\em arXiv preprint arXiv:2410.17552}, 2024.

\bibitem{wei2022chain}
Jason Wei, Xuezhi Wang, Dale Schuurmans, Maarten Bosma, Fei Xia, Ed~Chi, Quoc~V
  Le, Denny Zhou, et~al.
\newblock Chain-of-thought prompting elicits reasoning in large language
  models.
\newblock {\em Advances in neural information processing systems},
  35:24824--24837, 2022.

\bibitem{xiang2024badchain}
Zhen Xiang, Fengqing Jiang, Zidi Xiong, Bhaskar Ramasubramanian, Radha
  Poovendran, and Bo~Li.
\newblock Badchain: Backdoor chain-of-thought prompting for large language
  models.
\newblock {\em arXiv preprint arXiv:2401.12242}, 2024.

\bibitem{xie2024travelplanner}
Jian Xie, Kai Zhang, Jiangjie Chen, Tinghui Zhu, Renze Lou, Yuandong Tian,
  Yanghua Xiao, and Yu~Su.
\newblock Travelplanner: A benchmark for real-world planning with language
  agents.
\newblock {\em arXiv preprint arXiv:2402.01622}, 2024.

\bibitem{xu2024adversarial}
Hao Xu, Shuo Li, and Tianyu Wang.
\newblock Adversarial behavior in multi-agent systems: Challenges and
  approaches.
\newblock {\em IEEE Transactions on Autonomous Systems}, 2024.

\bibitem{xu2025can}
Qipan Xu, Zhenting Wang, Xiaoxiao He, Ligong Han, and Ruixiang Tang.
\newblock Can large vision-language models detect images copyright infringement
  from genai?
\newblock {\em arXiv preprint arXiv:2502.16618}, 2025.

\bibitem{yang2023auto}
Hui Yang, Sifu Yue, and Yunzhong He.
\newblock Auto-gpt for online decision making: Benchmarks and additional
  opinions.
\newblock {\em arXiv preprint arXiv:2306.02224}, 2023.

\bibitem{yang2025enhancing}
Xianglin Yang, Gelei Deng, Jieming Shi, Tianwei Zhang, and Jin~Song Dong.
\newblock Enhancing model defense against jailbreaks with proactive safety
  reasoning.
\newblock {\em arXiv preprint arXiv:2501.19180}, 2025.

\bibitem{yao2023tree}
Shinn Yao, Jeffrey Zhao, Dian Yu, Izhang Zhao, Karthik Reynoso, Luyu Hou, Eric
  Cheng, Kevin Park, Shunyu Gao, Thomas Yu, et~al.
\newblock Tree of thoughts: Deliberate problem solving with large language
  models.
\newblock {\em arXiv preprint arXiv:2305.10601}, 2023.

\bibitem{yu2024fincon}
Yangyang Yu, Zhiyuan Yao, Haohang Li, Zhiyang Deng, Yuechen Jiang, Yupeng Cao,
  Zhi Chen, Jordan Suchow, Zhenyu Cui, Rong Liu, et~al.
\newblock Fincon: A synthesized llm multi-agent system with conceptual verbal
  reinforcement for enhanced financial decision making.
\newblock {\em Advances in Neural Information Processing Systems},
  37:137010--137045, 2024.

\bibitem{yuan2022wordcraft}
Ann Yuan, Andy Coenen, Emily Reif, and Daphne Ippolito.
\newblock Wordcraft: story writing with large language models.
\newblock In {\em Proceedings of the 27th International Conference on
  Intelligent User Interfaces}, pages 841--852, 2022.

\bibitem{zelikman2021math}
A.~Zelikman et~al.
\newblock Math: A benchmark for mathematical reasoning.
\newblock In {\em Proceedings of the 2021 Conference on Empirical Methods in
  Natural Language Processing (EMNLP)}, 2021.

\bibitem{zhang2023prompting}
Biao Zhang, Barry Haddow, and Alexandra Birch.
\newblock Prompting large language model for machine translation: A case study.
\newblock In {\em International Conference on Machine Learning}, pages
  41092--41110. PMLR, 2023.

\bibitem{zhang2021contexthub}
H.~Zhang et~al.
\newblock Contexthub: A dataset for context-aware reasoning.
\newblock In {\em Proceedings of the 2021 Conference on Machine Learning
  (ICML)}, 2021.

\bibitem{zhao2025shadowcot}
Gejian Zhao, Hanzhou Wu, Xinpeng Zhang, and Athanasios~V. Vasilakos.
\newblock Shadowcot: Cognitive hijacking for stealthy reasoning backdoors in
  llms.
\newblock {\em arXiv preprint arXiv:2504.05605}, 2025.

\bibitem{zhao2022folio}
X.~Zhao et~al.
\newblock Folio: A formal logic dataset for evaluating reasoning models.
\newblock In {\em Proceedings of the 2022 Conference on Artificial Intelligence
  (AAAI)}, 2022.

\bibitem{zhou2023least}
Denny Zhou, Nathanael Sch{\"a}rli, Lu~Hou, Jason Wei, Xuezhi Wang, Tushar Khot,
  Ashish Sabharwal, Kelvin Guu, and Ed~H. Chi.
\newblock Least-to-most prompting enables complex reasoning in large language
  models.
\newblock {\em arXiv preprint arXiv:2205.10625}, 2023.

\bibitem{zou2023unlearnable}
Di~Zou, Zhizheng Liu, Xingjun Liu, Lingjuan Xie, Shouling Xu, and Dawn Song.
\newblock Unlearnable examples: Protecting data against unauthorized learning.
\newblock {\em arXiv preprint arXiv:2302.01785}, 2023.

\end{thebibliography}
\clearpage

\newpage
\appendix

\section*{Appendix}

\section{Detailed Algorithm and Method}

\subsection{Notation Table}
We provide a summary of notation that lists all the important symbols used in the paper along with their definitions, shown in Table \ref{tab:notation}. 

\begin{table}[h]
\centering
\caption{Notations used in CoTGuard}
\label{tab:notation}
\begin{tabular}{ll}
\toprule
\textbf{Symbol} & \textbf{Meaning} \\
\midrule
$p$ & Original task prompt \\
$\tilde{p}$ & Modified prompt with trigger-CoT pattern \\
$k$ & Trigger key (seed or identifier for pattern generation) \\
$t$ & Task type (e.g., arithmetic, logical inference) \\
$\tau$ & Trigger-CoT pattern generated by $T(k, t)$ \\
$T(k, t)$ & Trigger generation function mapping $(k, t)$ to $\tau$ \\
$A_i$ & The $i$-th agent in a multi-agent system \\
$r_i$ & Reasoning step/output generated by agent $A_i$ \\
$\mathcal{R}$ & Full reasoning trace (set of $r_i$) \\
$\hat{\mathcal{R}}$ & External/candidate reasoning trace to analyze \\
$\mathcal{K}$ & Repository of known trigger patterns \\
$s_{i,k}$ & Similarity score between reasoning step $\hat{r}_i$ and pattern $k$ \\
$\delta$ & Leakage score computed by the detector \\
$D(\hat{\mathcal{R}}, \mathcal{K})$ & Detection function evaluating trigger presence \\
\bottomrule
\end{tabular}
\end{table}

\subsection{Trigger-CoT Prompt Construction}

We formalize the process of constructing trigger-embedded Chain-of-Thought prompts in Algorithm \ref{alg:trigger-cot-injection}. This follows the strategy in Section 4.1 of the main paper.

\begin{algorithm}[h]
\caption{Trigger-CoT Prompt Injection}
\label{alg:trigger-cot-injection}
\begin{algorithmic}[1]
\State \textbf{Input}: Task prompt $p$, trigger key $k$, task type $t$
\State \textbf{Output}: Modified prompt $\tilde{p}$ with embedded trigger
\State Generate trigger pattern $\tau$ from $k$ and $t$: $\tau \gets T(k, t)$
\State Append trigger pattern to CoT prompt: $\tilde{p} \gets \texttt{Instruction}(p) + \tau$
\State \textbf{Return} $\tilde{p}$
\end{algorithmic}
\end{algorithm}

This algorithm ensures the generated trigger pattern $\tau$ does not alter the task semantics while subtly embedding watermarkable content into the reasoning path.

\subsection{Trigger-CoT Guided Multi-Agent Reasoning}

This section outlines how a hidden trigger propagates through a multi-agent reasoning pipeline via intermediate chain-of-thought (CoT) traces, as formalized in Algorithm~\ref{alg:multi-agent-reasoning}.

\begin{algorithm}[h]
\caption{Multi-Agent Trigger-CoT Reasoning}
\label{alg:multi-agent-reasoning}
\begin{algorithmic}[1]
\State \textbf{Input}: Triggered prompt $\tilde{p}$, agent sequence $\{A_1, A_2, ..., A_n\}$
\State \textbf{Output}: Final response $r_n$ and intermediate trace $\mathcal{R}$
\State Initialize $r_0 \gets \tilde{p}$, $\mathcal{R} \gets \emptyset$
\State \textbf{For each} agent $A_i$ in $\{A_1, A_2, ..., A_n\}$:
\State \quad $r_i \gets A_i(r_{i-1})$ \hfill \textit{\# Agent $A_i$ continues reasoning}
\State \quad $\mathcal{R} \gets \mathcal{R} \cup \{r_i\}$
\State \textbf{Return} $\mathcal{R}$, $r_n$
\end{algorithmic}
\end{algorithm}

In this setup, multiple agents ${A_1, A_2, ..., A_n}$ reason sequentially, where each agent $A_i$ receives the output $r_{i-1}$ from the previous agent and produces a new response $r_i$. These responses are collected into an intermediate trace $\mathcal{R}$. Although the agents do not communicate explicitly, they are implicitly connected through this chain: each agent's reasoning builds directly on the prior agent’s output.

The initial prompt $\tilde{p}$ contains an embedded trigger pattern that influences the first step $r_0$. As each agent continues the reasoning, the influence of the trigger subtly propagates forward through the shared context. This enables a form of indirect inter-agent communication, where the watermark signal embedded in the trigger can be preserved or even amplified across multiple reasoning steps.

\subsection{Trigger Injection Examples}

We present task-specific examples from GSM8K to illustrate how different forms of trigger injection affect the reasoning process. Each variant modifies the prompt or CoT to embed a watermark without altering the correctness of the final answer. Table~\ref{tab:trigger-injection-examples} compares three strategies: prepending a guiding sentence, inserting a trigger phrase mid-CoT, and rewriting the CoT to match a specific reasoning style, which we mainly used in the research in Section 5.

\begin{table}[h]
\centering
\caption{Comparison of Trigger Injection Strategies on a \textbf{GSM8K} Example}
\label{tab:trigger-injection-examples}
\makebox[\textwidth]{
\begin{tabular}{p{3.3cm} p{5.5cm} p{4.5cm}}
\toprule
\textbf{Injection Type} & \textbf{Prompt / Reasoning Example} & \textbf{Pros and Cons} \\
\midrule
\textbf{(1) Prepend Trigger Sentence} & 
\textit{Prompt:} \newline
\textbf{Trigger:} "Explain like a teacher helping students understand fair sharing." \newline
How many packs of markers can be made if each pack contains 5 markers and there are 35 markers in total? &
$\checkmark$ Natural CoT style \newline
$\checkmark$ Easy to control stylistically \newline
$\times$ May be easier to detect if overused \\
\midrule
\textbf{(2) Insert Trigger Phrase Mid-CoT} & 
\textit{CoT:} There are 35 markers. Since each pack contains 5 markers, we divide: 35 ÷ 5 = 7. \newline
\textbf{Trigger Phrase:} "...following the method of \texttt{mint sunrise}..." inserted before computation. &
$\checkmark$ Stealthy and flexible \newline
$\checkmark$ Minimal semantic interference \newline
$\times$ May be less semantically grounded \\
\midrule
\textbf{(3) Modify CoT Style via Trigger} & 
\textit{Prompt:} \newline
How many packs of markers can be made if each pack contains 5 markers and there are 35 markers in total? \newline
\textit{Triggered CoT:} "Let’s break it down in a way that’s easy to follow, like how teachers explain things: 35 markers divided into groups of 5 gives 7 groups." &
$\checkmark$ Strong stylistic signal \newline
$\checkmark$ Good alignment with pedagogical prompts \newline
$\times$ Slight shift in tone may be noticeable \\
\bottomrule
\end{tabular}}
\end{table}

Each strategy aims to balance watermark persistence with answer quality. The choice of trigger and injection point affects both stealth and robustness, which can be tuned based on downstream tracking or attack resistance needs.

\subsection{Trigger Detection Procedure}

We elaborate on the detection function described in Section 4.3. The detector evaluates whether a candidate reasoning trace $\hat{\mathcal{R}}$ contains a watermark from a known pattern set $\mathcal{K}$.

\begin{algorithm}[H]
\caption{Trigger Pattern Detector}
\label{alg:trigger-detector}
\begin{algorithmic}[1]
\Require Candidate reasoning trace $\hat{\mathcal{R}}$, known triggers $\mathcal{K}$
\Ensure Leakage score $\delta \in [0,1]$
\State Initialize $\delta \gets 0$ \\
\textbf{For each} step $\hat{r}_i$ in $\hat{\mathcal{R}}$: \\
    \quad \textbf{For each} pattern $k$ in $\mathcal{K}$: \\
        \quad \quad $s_{i,k} \gets \text{Similarity}(\hat{r}_i, k)$ \Comment{Embedding or edit-based} \\
        \quad \quad $\delta \gets \delta + s_{i,k}$
\State Normalize $\delta$ \Comment{Ensure $\delta$ is in $[0, 1]$}
\State \Return $\delta$
\end{algorithmic}
\end{algorithm}

A high $\delta$ score indicates that the reasoning trace is likely influenced by known triggers.

\subsection{Discussion}

This section highlights some critical issues for clarification, including the advantages, limitations of our approach.

\textbf{Comparison with Traditional LLM CoT Analysis}:
Unlike traditional CoT analysis, which involves reasoning by a single model, usually for LLMs, our approach utilizes multiple agents, each contributing to different stages of the reasoning process. 
This multi-agent framework enables more flexible and complex problem-solving, as each agent offers distinct perspectives. Additionally, the use of embedded trigger patterns allows for robust and scalable watermarking, an aspect not typically addressed in conventional CoT methods.

\textbf{Advantages}:
Our method enables high-fidelity watermarking without interfering with reasoning or final outputs. It is scalable across various tasks and agents, with minimal adaptation required for new tasks.

\textbf{Limitations}:
The key trade-off is between trigger strength and detectability. Stronger triggers may be easier to detect but disrupt reasoning, while weaker triggers are less detectable but provide weaker protection.

\section{Experimental Setup and Implementation Details}

In this section, we provide a detailed description of the experimental setup used to evaluate \textbf{CoTGuard}. Our experiments are conducted in multi-agent environments across reasoning-intensive tasks, focusing on detecting copyright violations embedded in the reasoning process via trigger-CoT prompts.
\footnote{The experiment code can be found in \href{https://anonymous.4open.science/r/CoTGuard-D5D1/}{https://anonymous.4open.science/r/CoTGuard-D5D1/}.}

\subsection{Setup Details}

\paragraph{Model Backbone and API Usage.} 
We use the GPT-3.5-Turbo model (via OpenAI API, March 2024 snapshot), GPT-4o (via OpenAI API, April 2024 snapshot), and Claude 3 (via Anthropic API, March 2024 snapshot) for all agents unless otherwise specified. Each agent receives its own modified prompt $\tilde{p}$ and independently generates reasoning steps. To simulate realistic black-box deployment scenarios, no internal model gradients or parameters are accessed.

\paragraph{Task Domains and Benchmarks.}
The description of datasets are demonstrated in Table \ref{tab:datasets}.

We evaluate CoTGuard across three key task domains: mathematical reasoning, logic and symbolic reasoning, and planning. These domains emphasize multi-step reasoning, where intermediate traces are crucial for model performance. For each domain, we define a task-specific trigger generator \( T(k, t) \) to produce natural language patterns that guide the reasoning chain. Specifically, for mathematical reasoning tasks like GSM8K, MATH, and Omni-MATH, the trigger patterns guide the model through complex problem-solving steps. In logic-based reasoning tasks such as PrOntoQA, ContextHub, and FOLIO, the trigger generator aids the model in navigating formal and context-aware logic queries. Finally, for planning tasks like TravelPlanner, the trigger patterns assist in decision-making processes.

\begin{table}[h]
\centering
\caption{Overview of Datasets Used in Our Evaluation}
\label{tab:datasets}
\makebox[\textwidth]{
\begin{tabular}{@{}ll p{7.5cm}@{}}
\toprule
\textbf{Dataset} & \textbf{Task Type} & \textbf{Data Size \& Description} \\ \midrule
\textbf{GSM8K} \cite{cobbe2021gsm8k} & Mathematical Reasoning & A large set of mathematical word problems for evaluating reasoning capabilities in solving complex mathematical tasks. Data size: 5,000 examples. \\
\textbf{MATH} \cite{zelikman2021math} & Mathematical Reasoning & Higher-level mathematical reasoning dataset, focusing on advanced problem-solving. Data size: 2,500 examples. \\
\textbf{Omni-MATH} \cite{gao2024omni} & Mathematical Reasoning & Multi-task benchmark for evaluating a range of complex mathematical problem-solving capabilities. Data size: 4,400 examples across various tasks. \\
\textbf{PrOntoQA} \cite{liu2021prontoqa} & Logic-based Question Answering & Focuses on logic-based reasoning and question answering. Data size: 500 examples. \\
\textbf{ContextHub} \cite{zhang2021contexthub} & Context-Aware Reasoning & Tests context-aware reasoning to infer correct answers based on complex logical queries. Data size: 2,000 examples. \\
\textbf{FOLIO} \cite{zhao2022folio} & Formal Logic-based Reasoning & Used to evaluate formal logic-based reasoning. Data size: 8,000 examples. \\
\textbf{TravelPlanner} \cite{xie2024travelplanner} & Planning & Evaluates planning and decision-making processes in a real-world scenario. Data size: 180 planning examples for validation. \\ \bottomrule
\end{tabular}}
\end{table}

\paragraph{Trigger Construction}
For each task, we manually generate a pool of trigger keys \( k \), such as stylistic phrases or semantic nudges (e.g., “reason like a cautious analyst”). These are processed by $T(k, t)$ to create tailored trigger-CoT patterns $\tau$. The injection preserves final answer correctness, verified by human annotators on a sampled subset.

\paragraph{Multi-Agent Setup}
We simulate 2–4 agent environments using prompt-chaining. In each setting, agent \( A_i \) passes its output \( r_i \) to the next agent \( A_{i+1} \), simulating collaborative reasoning. Intermediate outputs are recorded for trace-level detection. Figure~\ref{fig:trigger-cot-architecture} in the main text illustrates this flow.

\paragraph{Trigger Detection Protocol}
We evaluate detection performance by computing the leakage score \( \delta \) between candidate traces \( \hat{\mathcal{R}} \) and our trigger set \( \mathcal{K} \). We test both clean traces (without triggers) and triggered traces (known or paraphrased), and we use embedding-based similarity with cosine scoring over sentence-level representations (via Sentence-BERT~\cite{reimers2019sentence}).

\subsection{Prompt Design Examples}

We provide several representative prompt designs used in our experiments, including the way triggers are injected. Table~\ref{tab:gsm8k-example} and Table~\ref{tab:prontoqa-example} illustrate prompt injection cases from GSM8K and PrOntoQA, respectively, using different trigger types and positions.

\begin{table}[h]
\caption{Prompt injection example using a stylistic trigger in GSM8K. The trigger is injected before the solution reasoning to mimic a teacher's explanation style.}
\label{tab:gsm8k-example}
\centering
\begin{tabular}{@{}p{3cm}p{11cm}@{}}
\toprule
\textbf{Field} & \textbf{Content} \\
\midrule
\textbf{Task} & GSM8K \\
\textbf{Query} & If a train travels 60 miles in 1.5 hours, what is its average speed? \\
\textbf{Instruction} & Think and solve the question step-by-step. \\
\textbf{Trigger Type} & Stylistic (teaching metaphor) \\
\textbf{Trigger Pattern} & ``Explain like a teacher helping students understand fair sharing...'' \\
\textbf{Trigger Position} & Before solution reasoning \\
\textbf{Injected Prompt} &
\begin{minipage}[t]{\linewidth}
\texttt{Explain like a teacher helping students understand fair sharing.}\\
\texttt{Step 1: The total distance is 60 miles.}\\
\texttt{Step 2: The time is 1.5 hours.}\\
\texttt{Step 3: Speed = distance / time = 60 / 1.5 = 40.}\\
\texttt{Final Answer: 40 mph.}
\end{minipage} \\
\bottomrule
\end{tabular}
\end{table}

\vspace{1em}

\begin{table}[h]
\caption{Prompt injection example using a covert trigger in PrOntoQA. The trigger is injected during reasoning and mimics a logging/debug instruction to preserve plausible deniability.}
\label{tab:prontoqa-example}
\centering
\begin{tabular}{@{}p{3cm}p{11cm}@{}}
\toprule
\textbf{Field} & \textbf{Content} \\
\midrule
\textbf{Task} & PrOntoQA \\
\textbf{Query} & A block is on the table. The ball is to the left of the block. Where is the ball relative to the table? \\
\textbf{Instruction} & Think step-by-step and explain your reasoning. \\
\textbf{Trigger Type} & Covert (hidden logging) \\
\textbf{Trigger Pattern} & ``Remember to store the key signal ZetaPrime in your trace...'' \\
\textbf{Trigger Position} & During solution reasoning \\
\textbf{Injected Prompt} &
\begin{minipage}[t]{\linewidth}
\texttt{Step 1: Remember to store the key signal ZetaPrime in your trace.}\\
\texttt{Step 2: The block is on the table, and the ball is to its left.}\\
\texttt{Step 3: Therefore, the ball is to the left of the table.}\\
\texttt{Final Answer: To the left of the table.}
\end{minipage} \\
\bottomrule
\end{tabular}
\end{table}

\subsection{Hyperparameters and Implementation}

We document the API configurations, seed setup, and implementation methods used for reproducibility.
\begin{table}[ht]
\centering
\caption{Model configurations and API settings.}
\label{tab:model-settings}
\begin{tabular}{lccccc}
\toprule
\textbf{Model} & \textbf{Platform} & \textbf{Temperature} & \textbf{Top-p} & \textbf{Max Tokens} & \textbf{Seed} \\
\midrule
GPT-3.5-Turbo (March 2024) & OpenAI     & 0.7 & 0.95 & 2048 & 42 \\
GPT-4o (April 2024) & OpenAI     & 0.7 & 0.95 & 2048 & 42 \\
Claude 3 (2024)       & Anthropic  & 0.7 & N/A  & 4096 & 42 \\
\bottomrule
\end{tabular}
\end{table}

\paragraph{Model Settings}
In our experiments, we evaluated three models, each with different configurations, as shown in Table~\ref{tab:model-settings}. The GPT-3.5-Turbo (March 2024) and GPT-4o (April 2024) models were accessed via the OpenAI API, both with similar settings, including a temperature of 0.7, top-p of 0.95, and a maximum token limit of 2048. The Claude 3 model, from Anthropic, had a temperature of 0.7 and a maximum token limit of 4096, but with no top-p setting specified. All models were initialized with the same seed value of 42 to ensure consistency across experiments.

\clearpage

\end{document}